\documentclass[sigconf, nonacm]{acmart}

\AtBeginDocument{%
  \providecommand\BibTeX{{%
    \normalfont B\kern-0.5em{\scshape i\kern-0.25em b}\kern-0.8em\TeX}}}

\usepackage{multirow}
\usepackage{bbm}
\usepackage{algorithm}
\usepackage{algorithmic}
\input{Definitions}
\def\Tau{{\mathcal{T}}}
\def\clip{\textsf{clip}}

\graphicspath{{figures/}}

\begin{document}

\title{Towards Understanding the Adversarial Vulnerability of Skeleton-based Action Recognition}

\author{Tianhang Zheng$^{1}$, Sheng Liu$^{2}$, Changyou Chen$^{2}$, Junsong Yuan$^{2}$, Baochun Li$^{1}$, Kui Ren$^{3}$}
\affiliation{\institution{$^{1}$University of Toronto, $^{2}$State University of New York at Buffalo, $^{3}$Zhejiang University}}

\renewcommand{\shortauthors}{}

\begin{abstract}
Skeleton-based action recognition has attracted increasing attention due to its potentially broad applications such as autonomous and anonymous surveillance. With the help of deep learning techniques, it has also witnessed substantial progress and achieved excellent accuracy in non-adversarial environments. However, in practice, potential adversaries might easily deceive an action recognition model by performing actions with imperceptible perturbations. Deploying such a model without understanding its adversarial vulnerability might lead to severe consequences, {\em e.g.}, recognizing a violent action as a normal one. Despite these security concerns, research on the vulnerability of skeleton-based action recognition remains scant, partly due to the challenges caused by the unique nature of human skeletons and actions. Specifically, we argue that for imperceptible and reproducible adversarial skeleton actions: 1) the bone lengths should be maintained roughly the same as the original bone lengths; 2) the changes of joint angles should be small; 3) the adversarial motion speeds should be restricted. These unique constraints hinder direct applications of existing attack methods to adversarial skeleton actions.

In this paper, we conduct a thorough study towards understanding the adversarial vulnerability of skeleton-based action recognition. We first formulate the generation of adversarial skeleton actions as a constrained optimization problem by representing or approximating the constraints with mathematical equations. To deal with the intractable primal optimization problem with equality constraints, we propose to optimize its unconstrained dual problem using ADMM. We further design an efficient plug-in defense, inspired by recent theories and empirical observations, against adversarial skeleton actions. Extensive evaluations demonstrate the effectiveness of our attack and defense, and reveal the properties of adversarial skeleton actions.
\end{abstract}



\maketitle

\section{Introduction}
Action recognition is an important task in multimedia and computer vision, motivated by many downstream applications such as video surveillance and indexing, and human-machine interaction \cite{cheng2015advances}. It is also a challenging task since it requires capturing long-term spatial-temporal motion patterns to understand the semantics of actions. Many recent works from the multimedia \& computer vision community \cite{wang2016action, ke2017a, gao2019optimized, shi2019two, shi2019skeleton, yang2020hierarchical} propose to learn action recognition on the human skeleton motion captured by cameras or depth sensors, where an action is represented by a time series of poses represented as 3D body skeletons. Compared with video streams, skeleton representation is more robust to the variance of background clutters, and also easier-to-handle for machine learning models due to its compact representation. Recent advances in deep learning techniques have been applied to skeleton-based action recognition, including convolutional neural networks \cite{ke2017a, li2018co}, recurrent neural networks \cite{li2018CVPR, si2019attention}, and graph neural networks \cite{yan2018spatial, shi2019two, gao2019optimized, liu2020disentangling}. 
\begin{figure}
    \centering
    \includegraphics[width=0.95\linewidth]{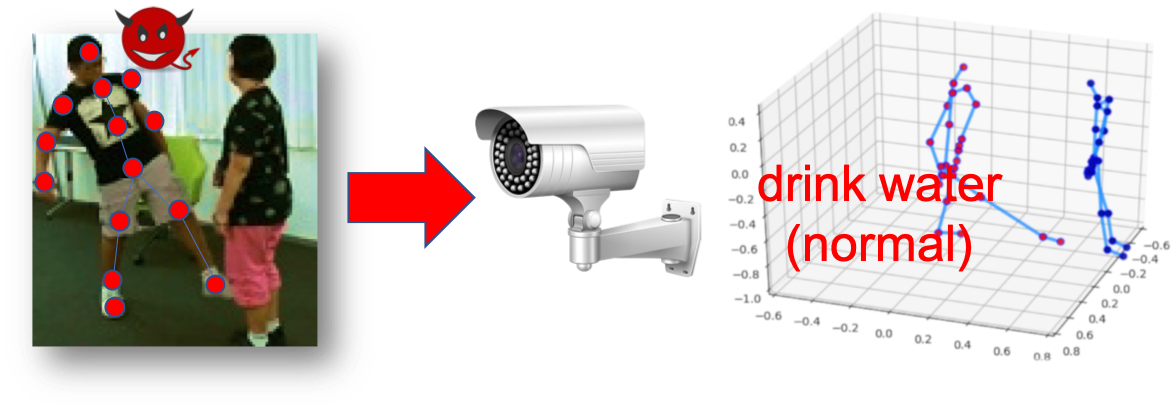}
    \caption{The targeted setting: misleading the model to recognize ``kicking person" as ``drinking water" (normal action) by perturbing the skeleton action. To launch the attack in a real-world scenario ({\em e.g.,} under a surveillance camera), the adversarial skeleton action should satisfy certain constraints. The figure is drawn based on \cite{shahroudy2016ntu}.}
    \vspace{-0.3cm}
    \label{fig:target_diagram}
\end{figure}
On the other hand, existing work has demonstrated the vulnerability of deep learning techniques to adversarial examples in many application domains, such as face recognition and object detection. 
This phenomenon motivates us to suspect that, despite achieving high accuracy in non-adversarial environments, the deep neural networks (DNNs) for skeleton-based action recognition might also be vulnerable to adversarial skeleton actions. 

It is worth noting that a thorough study on the adversarial vulnerability of action-recognition models is indispensable before deploying them to real-world applications such as surveillance systems. Otherwise, the potential adversaries might easily deceive those systems by performing specific adversarial actions, leading to significant consequences, as shown in Fig.~\ref{fig:target_diagram}.
To our knowledge, the study on adversarial skeleton actions is scant and non-trivial\footnote{The only parallel work is detailed in section~\ref{subsec:adv_attack}.}, due to the fundamental differences between the properties of adversarial skeleton actions and other adversarial examples. 
The differences are caused by the bones between joints and the joint angles, which impose unique spatial constraints on skeleton data \cite{shahroudy2016ntu}. Specifically, in the generated adversarial skeleton actions, lengths of bones must be maintained the same, and simultaneously, joint angles cannot violate certain physiological structures. In addition, considering the physical properties of human bodies, the speeds of motions in the adversarial actions should also be constrained. {\em If any constraint is not satisfied, the adversarial skeleton actions might be easily perceived and detected or could not be performed by the actors.} 

To understand the adversarial vulnerability of skeleton-based action recognition, we first study how to generate adversarial skeleton actions. Specifically, we formulate the generation of adversarial skeleton actions as a constrained optimization problem by representing the spatio-temporal constraints with mathematical equations. Since the primal constrained problem is intractable, we turn to solve its dual problem. Moreover, since all the constraints are represented by mathematical equations, both primal and dual variables become unconstrained in the dual problem. We further specify an efficient algorithm based on ADMM to solve the unconstrained dual problem, in which the internal minimization objective is optimized by an Adam optimizer, and the external maximization objective is optimized by one-step gradient ascent. We show that this algorithm can find an adversarial skeleton action within a couple of hundred internal steps. 

Other than the attack, we further propose an efficient defense against adversarial skeleton actions based on recent theories and empirical observations. Our defense consists of two core steps, {\em i.e.,} adding Gaussian noise and Gaussian filtering to action data. The first step, adding Gaussian noise, is inspired by the recent advance in certified defenses. Specifically, adding Gaussian noise to the input is proved to be a certified defense, which means additive Gaussian noise on the adversarial examples can guarantee the model to output a correct prediction (with high probability), as long as the adversarial perturbation is restricted within a certain radius in the neighbor of the original data sample. Note that there are several other methods to certify model robustness, such as dual approach, interval analysis, and abstract interpretations \cite{dvijotham2018dual, wong2018provable, mirman2018differentiable, gowal2018effectiveness, wang2018efficient}. We adopt the Gaussian noise method because it is simple, effective, and more importantly, scalable to complicated models. Note that skeleton-based action recognition models are always more complicated than the common ConvNets certified by \cite{dvijotham2018dual, wong2018provable, mirman2018differentiable, gowal2018effectiveness, wang2018efficient}. The second step is to smooth the skeleton frames along the temporal axis using a Gaussian filter. This step will not affect the robustness certified by the first step according to the post-processing property \cite{lecuyer2018certified, li2018second, cohen2019certified}, but can always filter out a certain amount of adversarial perturbation and random noise in practice, thus making our defense applicable to normally trained models. 

Our proposed attack and defense are evaluated on two opensource models, {\it i.e.}, 2s-AGCN and HCN \cite{shi2019two, li2018co} \footnote{We select these two models because the authors have released the code and hyperparameters on Github so that we can correctly reproduce the results. Also, these two models achieve fairly good performance.}. Extensive evaluations show that our attack can achieve $100\%$ attack success rate with almost no violation of the constraints. Moreover, the visualization results, including images and videos, demonstrate that the difference between the original and adversarial skeleton actions is imperceptible. Extensive evaluations also show that our defense is effective and efficient. Specifically, our defense can improve the empirical accuracy of normally trained models to over $60\%$ against adversarial skeleton actions under different settings.

To summarize, our main contribution is three-fold:
\begin{enumerate}
    \item We identify the constraints needed to be considered in adversarial skeleton actions, and formulate the problem of generating adversarial skeleton actions as a constrained optimization problem by formulating those constraints as mathematical equations. We further propose to solve the primal constrained problem by optimizing its dual problem using ADMM, achieving 100\% attack success rate.
    \item We propose an efficient two-step defense against adversarial skeleton actions based on previous theories and empirical observations, and specify the defense in both inference and certification stages. The proposed defense achieves high robust accuracy under mild perturbations.
    \item We conduct extensive evaluations on two opensource models and two datasets. We also provide several interesting observations regarding the properties of adversarial skeleton actions based on the experimental results. 
\end{enumerate}


\section{Preliminaries}
\subsection{Definitions and Notations}
\begin{figure}
    \centering
    \includegraphics[width=0.36\linewidth]{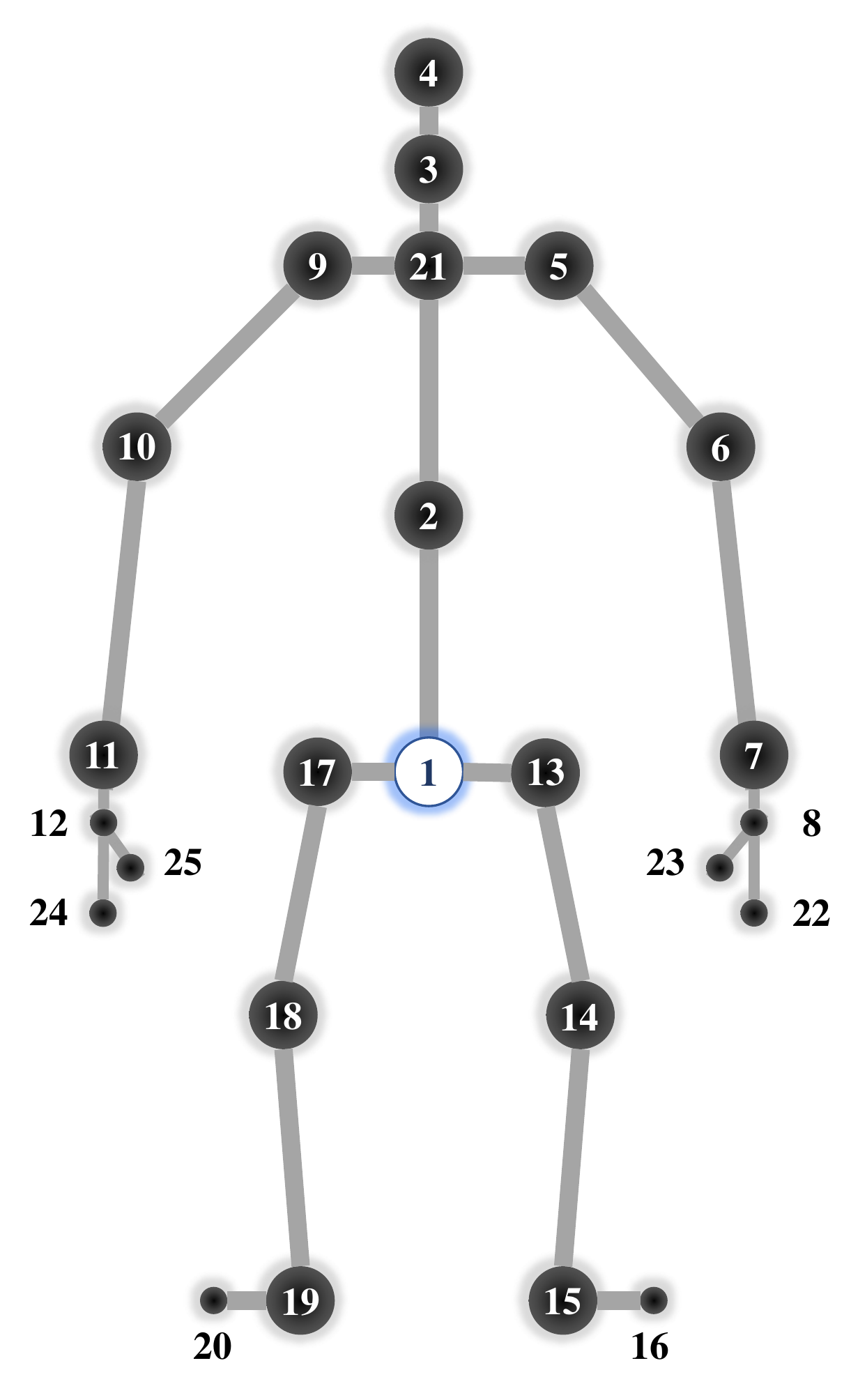}
    \hspace{0.35cm}
    \includegraphics[width=0.34\linewidth]{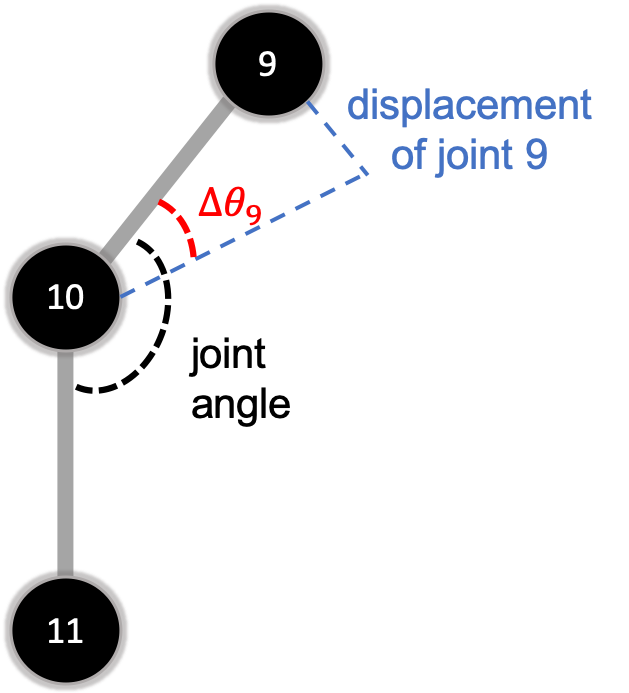}
    \caption{Skeleton Representation}
    \vspace{-0.3cm}
    \label{fig:skeleton}
\end{figure}

Let $\xb$ and $l \in \{1, 2,...,L\}$ respectively denote a data sample and the label, where $L$ is the number of all possible classes. For an image, $\xb$ is a 2D matrix. For a skeleton action studied in this paper, $\xb \triangleq \{(x^\tau_i, y^\tau_i, z^\tau_i)_{i=1}^I\}_{\tau=1}^\Tau$, where $(x^\tau_i, y^\tau_i, z^\tau_i)$ denotes the position (coordinates) of the $i$-th joint of the $\tau$-th skeleton frame in an action sequence, with $I$ and $\Tau$ denoting the number of joints in a skeleton and the number of skeleton frames in an action sequence, respectively. The corresponding adversarial skeleton action is denoted by $\xb' \triangleq \{(x'^\tau_i, y'^\tau_i, z'^\tau_i)_{i=1}^I\}_{\tau=1}^\Tau$. We take the skeletons in the largest dataset, \textit{i.e.}, NTU RGB+D dataset, as an example. As shown in Fig.~\ref{fig:skeleton}, in a skeleton, there are totally 25 joints in a skeleton frame, and thus $I=25$. The number of frames $\Tau$ differs for each skeleton action, and usually, we subsample a constant number of frames from each sequence or pad zeros after each sequence to endow all the skeleton actions with the same $\Tau$. Let $\Fb_\Thetab(\cdot)$ denote a classification network, where $\Thetab$ represents the network weights. The logit output on $\xb$ is denoted by $\Fb_\Thetab(\xb)$ with $L$ elements ($\{\Fb_{\Thetab, k}(\xb)~|~k=1,...,L\}$). $\Fb_\Thetab(\cdot)$ can correctly classify $\xb$ iff $\argmax_{k} \Fb_{\Thetab, k}(\mathbf{x}) = l$. The goal of adversarial attacks is to find an adversarial sample $\xb'$, which satisfies several pre-defined constraints, such that $\argmax_{k} \Fb_{\Thetab, k}(\xb') \neq l$ or $\argmax_{k} \Fb_{\Thetab, k}(\mathbf{x}) = l_t$ ($l_t$ is the target label). A commonly-used constraint is that $\xb'$ should be close to the original sample $\xb$ according to some distance metric. 
\subsection{DNNs for Skeleton-based Action Recognition}
In the following, we briefly introduce the two DNNs used for evaluation of our proposed attack method in this project.
HCN is a CNN-based end-to-end hierarchical network for learning global co-occurrence features from skeleton data \cite{li2018co}. 
HCN is designed to learn different levels of features from both raw skeleton and skeleton motion. The joint-level features are learned by a multi-layer CNN, and the global co-occurrence features are learned from the fused joint-level features. At the end, the co-occurrence features are also fed to a fully-connected network for action classification.
2s-AGCN is one of the state-of-the-art GCN-based models for skeleton-based action recognition. In contrast to the earliest GCN-based model, ({\em i.e.,} ST-GCN), 2s-AGCN learns the appropriate graph topology of every skeleton action rather than prefine the graph topology. This enables 2s-AGCN to capture the implicit connections between joints in certain actions, such as the connection between hand and face in the ``wiping face" action. Besides, 2s-AGCN also adopts the two-stream framework to learn from both static and motion information. Overall, 2s-AGCN significantly improves the accuracy of ST-GCN by nearly 7\%. 

\subsection{Adversarial Attacks}\label{subsec:adv_attack}
After the discovery of adversarial examples, the community has developed hundreds of attack methods to generate adversarial samples. In the following, we mainly introduce four attack methods plus a parallel work, with a discussion on the difference between our proposed method and these attacks.
\paragraph{Fast Gradient Sign Method (FGSM)} FGSM is a typical one-step adversarial attack algorithm proposed by \cite{goodfellow2014}. The algorithm updates a benign sample along the direction of the gradient of the loss w.r.t. the sample. Formally, FGSM follows the update rule as 
\begin{equation}
\xb^\prime = \clip_{[v_{min}, v_{max}]}\{\xb + \epsilon \cdot \sign(\nabla_{\xb} \mathcal{L}(\Thetab, \xb, l))\}~,
\end{equation}\label{eq:fgsm}
where $\epsilon$ controls the maximum $\ell_\infty$ perturbation of the adversarial samples; $[v_{min}, v_{max}]$ is the valid element-wise value range and $\clip_{[a,b]}(\cdot)$ function clips its input into the range of $[a, b]$.

\paragraph{Projected Gradient Descent (PGD)} PGD \cite{kurakin2016adversarial, madry2017towards} is a strong iterative version of FGSM, which executes Eq. \ref{eq:fgsm} for multiple steps with a smaller step size and then projects the updated adversarial examples into the pre-defined $\ell_p$-norm ball. Specifically, in each step, PGD updates the sample by
\begin{align}
\xb^\prime_{t+1} = Proj\{\xb^\prime_{t}  + \alpha \cdot \sign(\nabla_{\xb^\prime_{t}} \mathcal{L}(\Thetab, \xb^\prime_{t}, l)\}~
\end{align}
The $Proj$ function is a clip function for $\ell_\infty$-norm balls, and an $\ell_2$ normalizer for $\ell_\infty$-norm balls.
\paragraph{Carlini and Wagner Attack}
\cite{carlini2017towards} proposes an attack called C\&W attack, which generates $\ell_p$-norm adversarial samples by optimization over the C\&W loss: 
\begin{align}\label{eq:cw_adv_optimization}
\min_{\xb'}D(\xb, \xb') + c \cdot loss(\xb')~.
\end{align}
In the C\&W loss, $D(\xb, \xb')$ represents some distance metric between the benign sample $\xb$ and the adversarial sample $\xb'$, and the metrics used in \cite{carlini2017towards} include $\ell_\infty$, $\ell_0$, and $\ell_2$ distances. $loss(\cdot)$ is a customized loss. It is worth noting that our proposed attack is completely different from PGD or C\&W attack. For PGD, C\&W, or many other attacks, the simple constraints on the pixel value can be resolved by projection functions or naturally incorporated into the objective by $sigmoid$/$tanh$ function. However, in our scenario, the constrained optimization problem is much more complicated, and thus has to be solved by more advanced methods.
\paragraph{ADMM-based Attack} 
\cite{zhao2018admm} also proposes a framework based on ADMM to generate $\ell_p$ adversarial examples. However, we note that our proposed attack is completely different from theirs in two aspects: First, the constraints we consider in this paper are much more complicated than the $\ell_p$-norm constraints in \cite{zhao2018admm}. Second, we formulate the problem in a very different manner. Specifically, \cite{zhao2018admm} follows the ADMM framework to break the problem defined like Eq.~\ref{eq:cw_adv_optimization} into two sub-problems; while our attack formulates a different problem with indispensable equality constraints, where ADMM is a natural solution to this problem.

\paragraph{Adversarial Attack on Skeleton Action}
Note that \cite{liu2019adversarial} is a parallel work that proposes an attack based on FGSM and BIM (PGD) to generate adversarial skeleton actions. Specifically, \cite{liu2019adversarial} adapts the FGSM and BIM to skeleton-based action recognition by using a clipping function and an alignment operation to impose the bone and joint constraints on the updated adversarial skeleton actions in each iteration. {\em However, \cite{liu2019adversarial} is very different from our work.} First, the joint constraints considered in \cite{liu2019adversarial} are not the constraints for joint angles mentioned before. 
Second, the alignment operation might corrupt the perturbation learned in each iteration. In contrast to \cite{liu2019adversarial}, we attempt to formulate adversarial skeleton action generation as a constrained optimization problem with equality constraints. Releasing the equality constraints by Lagrangian multipliers yields an unconstrained dual optimization problem, which does not need any complicated additional operation in the optimization process. Third, we propose to solve the the dual optimization problem by ADMM, which is a more appropriate method to optimize complicated constrained problems. Therefore, the attack achieves better performance than \cite{liu2019adversarial}, which will be detailed in section~\ref{sec:attack_performance}. Finally, we specify a defense method against adversarial skeleton actions based on the state-of-the-art theories and our observations. 

\subsection{Alternating Direction Method of Multipliers (ADMM)}
Alternating Direction Method of Multipliers (ADMM) is a powerful optimization algorithm to handle large-scale statistical tasks in diverse application domains. It blends the decomposability of dual ascent with the great convergence property of the method of multipliers. Currently, ADMM plays a significant role in solving statistical problems, such as support vector machines \cite{forero2010consensus}, trace norm regularized least squares minimization \cite{yang2013fast}, and constrained sparse regression \cite{bioucas2010alternating}.
Except for convex problems, ADMM is also a widely used solution to some nonconvex problems, whose objective function could be nonconvex, nonsmooth, or both. \cite{wang2019global} shows that ADMM is able to converge as long as the objective has a smooth part, while the remaining part can be coupled or nonconvex, or include separable nonsmooth functions. Applications of ADMM to nonconvex problems include network reference \cite{miksik2014distributed}, global conformal mapping \cite{lai2014splitting}, noisy color image restoration \cite{lai2014splitting}.
\subsection{Adversarial Defenses}
Both learning and security communities have developed many defensive methods against adversarial examples. Among them, adversarial training and several certified defenses attract the most attention due to their outstanding/guaranteed performance against strong attacks \cite{he2017adversarial, uesato2018adversarial, athalye2018obfuscated}. In the following, we briefly introduce adversarial training and several certified defenses, including the randomized smoothing method adopted in this paper.  
\paragraph{Adversarial Training}
Adversarial training is one of the most successful empirical defenses in the past few years \cite{goodfellow2014, madry2017towards, zhang2019theoretically}. The intuition of adversarial training is to improve model robustness by training the model with adversarial examples. Although adversarial training achieves tremendous success against many strong attacks \cite{zheng2019distributionally, andriushchenko2019square, tashiro2020output}, its performance is not theoretically guaranteed and thus might be compromised in the future. Besides, adversarial training always requires much more computational resource than standard training, making it not scalable to complicated models. 
\paragraph{Certified Defenses}
A defense with a theoretical guarantee on its defensive performance is considered as a certified defense. In general, there are three main approaches to design certified defenses. The first approach is to formulate the certification problem as an optimization problem and bound it by dual approach and convex relaxations \cite{dvijotham2018dual, raghunathan2018certified, wong2018provable}. The second approach approximates a convex set that contains all the possible outputs of each layer to certify an upper bound on the range of the final output \cite{mirman2018differentiable, gowal2018effectiveness, wang2018efficient}. The third is the randomized smoothing method used in this paper. The only essential operation for this method is to add Gaussian/Laplace noise to the inputs, which is simple and applicable to any deep learning models. \cite{lecuyer2018certified} first proves that randomized smoothing is a certified defense by theories on differential privacy. \cite{li2018second} improves the certified bound using a lemma on Renyi divergence. Cohen et al. \cite{cohen2019certified} proves a tight bound on the $\ell_2$ robust radius certified by adding Gaussian noise using the Neyman-Pearson lemma. \cite{jia2019certified} further extends the approach of \cite{cohen2019certified} to the $top-k$ classification setting. 
Since the bound proved by \cite{cohen2019certified} is the tightest, the method in \cite{cohen2019certified} is used for certification. In this paper we adopt the approach in \cite{lecuyer2018certified} due to its ability for efficient inference in practice. 
\section{Threat Model}
\subsection{Adversary Knowledge: White-box Setting}
In this paper, we follow the white-box setting, where the adversary has full access to the model architecture and parameters. We make this assumption because (i)
it is always a safe, conservative, and realistic assumption since we might never know the knowledge of potential adversaries about the model \cite{carlini2017towards}, which varies among different adversaries and also changes over time. (ii) For systems/devices equipped with an action recognition model, recognition is more likely to be done locally, or on a local cloud, making the adversary easily acquire the model parameters with his own system/device. Note that although most of the experiments on the proposed attack and defense are done under the white-box setting, we also have several experiments on evaluating the transferability of our attack.


\subsection{Adversary Goal: Targeted \& Untargeted label Setting}
Under the targeted setting, the goal of an adversary is to mislead the recognition model to predict the adversarial skeleton action as a targeted label pre-defined by the adversary. For instance, suppose the adversary is ``kicking" someone under a surveillance camera equipped with an action recognition model. It may launch a targeted attack to mislead the model to recognize this violent action as a normal one such as "drinking water". Under the untargeted label settings, an adversary only aims to disable the recognition and thus is considered successful as long as the model makes wrong predictions instead of a specific targeted prediction. In this paper, we propose two objectives suitable for the above two settings respectively, which will be detailed in section~\ref{subsec:constrained}.

\subsection{Imperceptibility \& Reproducibility}
Except for the aforementioned adversary goals, the adversary also requires the adversarial perturbation to be both imperceptible and reproducible. Here ``imperceptibility" means it should be difficult for human vision to figure out the difference between the original and adversarial skeleton actions. Imperceptibility is not only a common requirement in the previous attacks, but also a useful one in our scenario. Note that it is natural to schedule a periodical examination for an autonomous surveillance system by human labor to check if the system works well. If the system has been fooled by a seemingly ``normal" adversarial skeleton action, the mistake might be considered due to the system itself rather than the adversary who performs the adversarial skeleton action in the examination process. Here ``reproducibility'' is an additional requirement specific to our scenario. As mentioned in the introduction, the adversarial skeleton action could be a real threat when it can be reproduced under a real-world system.
Thus, to make our attack a real-world threat, the generated adversarial skeleton actions should satisfy three concrete constraints to be imperceptible and reproducible, which will be detailed in section~\ref{sec:attack}. 

\section{Adversarial Skeleton Action}\label{sec:attack}
In this section, we present our proposed attack, {\em i.e.,} ADMM attack. We first introduce how to formulate the three constraints into mathematical equations. Then we formulate the constrained optimization problem to generate adversarial skeleton actions under both targeted and untargeted settings. Finally, we elaborate on how to solve the optimization problem by ADMM.
\subsection{Bone Constraints}
We again take the skeletons in the NTU RGB+D dataset as an example. As shown in Fig.~\ref{fig:skeleton}, in a skeleton, there are totally 25 joints, forming a total of 24 bones. 
While the bones are not explicitly considered in modeling, they are strictly connecting to the 25 joints, thus imposing 24 bone-length constraints, {\it \textit{i.e.}}, the distance between the joints at the two ends of a bone should remain the same in adversarial skeleton actions. To mathematically represent the 24 bones, we associate each joint with its preceding joint, forming the two ends of a bone. As a result, the 24 preceding-joints for joint-2$\sim$joint-25 are denoted as $\mathcal{P} \triangleq \{(x^\tau_{pi}, y^\tau_{pi}, z^\tau_{pi})_{i=2}^{25}\}$. The corresponding joint indices of the elements in $\mathcal{P}$ are \{1, 21, 3, 21, 5, 6, 7, 21, 9, 10, 11, 1, 13, 14, 15, 1, 17, 18, 19, 2, 8, 8, 12, 12\}. We define the $i$-th bone's length as $B^\tau_i \triangleq \sqrt{(x^\tau_i - x^\tau_{pi})^2 + (y^\tau_i - y^\tau_{pi})^2 + (z^\tau_i - z^\tau_{pi})^2}$. In this regard, the bone constraints can be represented as $B^\tau_i = B'^\tau_i$. Due to the measurement errors in the NTU dataset itself, here we also tolerate very small difference between $B^\tau_i$ and $B'^\tau_i$. Therefore, we can finally formulate the bone constraints as
\begin{equation}\label{eq:con1}
    |B'^\tau_i - B^\tau_i|/B^\tau_i \leq \epsilon_L,
\end{equation}
where $\epsilon_L$ is usually set as $0.01\sim0.03$. {\em Note that inequality constraints in the primal problem will impose inequality constraints on the corresponding Lagrangian variables in the dual problem.} In order to avoid this in the dual problem, we reformulate the above inequality constraints as mathematical equations, {\it i.e.}, \eqref{eq:con1} is equivalent to
\begin{equation}\label{eq:bone_constraints}
    \max\{|B'^\tau_i - B^\tau_i|/B^\tau_i - \epsilon_L, 0\} = 0.
\end{equation}
\subsection{Joint Angle Constraints}
Except for the bone-length constraints, we also need to impose constraints on the rotations of the joint angles according to the physiological structures of human beings. Let us also use the NTU dataset as an example. Each joint angle corresponds to the angle between two bones, and thus can be represented by the three joint locations of those two corresponding bones as illustrated in the right of Fig.~\ref{fig:skeleton}. Note that a natural way to compute the joint angle as shown in Fig.~\ref{fig:skeleton} is to first compute the cosine value and then input the value into the arccos function. However, the gradient of arccos function is likely exploded, causing large numerical errors when the $\cos$ value of the joint angle is close to $1$ ($\frac{d}{dx} arccos x = -\frac{1}{\sqrt{1 - x^2}}$). To deal with this issue, we derive an approximate upper bound for the changes of joint angle value to avoid computing the arccos function and its gradient. Again, take the right of Fig.~\ref{fig:skeleton} as an example, the angle change $\Delta\theta_9$ caused by the displacement of joint-9 ({\em i.e.,} $x'^\tau_{9} - x^\tau_{9}$, $y'^\tau_{9} - y^\tau_{9}$, $z'^\tau_{9} - z^\tau_{9}$) can be approximated by $\sin\Delta\theta_9 \approx \frac{\sqrt{(x'^\tau_{9} - x^\tau_{9})^2 + (y'^\tau_{9} - y^\tau_{9})^2 + (z'^\tau_{9} - z^\tau_{9})^2}}{\sqrt{(x^\tau_{10} - x^\tau_{9})^2 + (y^\tau_{10} - y^\tau_{9})^2 + (z^\tau_{10} - z^\tau_{9})^2}}$. In particular, when the angle change $\Delta\theta$ is smaller than $0.1$ ({\em i.e.,} $5.73^{\circ}$), we can consider $sin\Delta\theta$ almost same as $\Delta\theta$. The total angle change $\Delta\theta$ is upper bounded by the sum of the changes caused by the displacements of joint-9, joint-10, and joint-11. Therefore the upper bound can be represented by
$
    J'^\tau = \frac{\sqrt{(x'^\tau_{9} - x^\tau_{9})^2 + (y'^\tau_{9} - y^\tau_{9})^2 + (z'^\tau_{9} - z^\tau_{9})^2} }{\sqrt{(x^\tau_{10} - x^\tau_{9})^2 + (y^\tau_{10} - y^\tau_{9})^2 + (z^\tau_{10} - z^\tau_{9})^2}}
    + \\ \frac{\sqrt{(x'^\tau_{10} - x^\tau_{10})^2 + (y'^\tau_{10} - y^\tau_{10})^2 + (z'^\tau_{10} - z^\tau_{10})^2} }{\sqrt{(x^\tau_{10} - x^\tau_{9})^2 + (y^\tau_{10} - y^\tau_{9})^2 + (z^\tau_{10} - z^\tau_{9})^2}}
    + \frac{\sqrt{(x'^\tau_{10} - x^\tau_{10})^2 + (y'^\tau_{10} - y^\tau_{10})^2 + (z'^\tau_{10} - z^\tau_{10})^2} }{\sqrt{(x^\tau_{11} - x^\tau_{10})^2 + (y^\tau_{11} - y^\tau_{10})^2 + (z^\tau_{11} - z^\tau_{10})^2}} \\
    + \frac{\sqrt{(x'^\tau_{11} - x^\tau_{11})^2 + (y'^\tau_{11} - y^\tau_{11})^2 + (z'^\tau_{11} - z^\tau_{11})^2} }{\sqrt{(x^\tau_{11} - x^\tau_{10})^2 + (y^\tau_{11} - y^\tau_{10})^2 + (z^\tau_{11} - z^\tau_{10})^2}} 
$
Although this representation looks more complicated than the arccos function, its gradient can be computed efficiently and accurately. Given such an approximation, the joint angle constraints can be similarly represented as 
\begin{equation}\label{eq:joint_angle_constraint}
    \max\{J_k'^{\tau} - \epsilon_J, 0\} = 0
\end{equation}
where $\epsilon_J$ is set as $0.1\sim0.2$ ($6^\circ \sim 12^\circ$). {\em Note that $J_k'^{\tau}$ represents the approximation of the change of a joint angle.}

\subsection{Speed Constraints}
According to the physical conditions of human beings, we should consider one more type of constraints, {\em i.e.,} temporal smoothness constraints. By those constraints, we attempt to restrict the speeds of the motions in the generated adversarial skeleton actions. Specifically, the speeds of the motions can be approximated by the displacements between two consecutive temporal frames, {\em i.e.,} $S^\tau_m \approx \sqrt{(x^{\tau+1}_m - x^\tau_m)^2 + (y^{\tau+1}_m - y^\tau_m)^2 + (z^{\tau+1}_m - z^\tau_m)^2}$. Then, similar to Eq.~\ref{eq:bone_constraints}, we bound the change of speeds by
\begin{equation}\label{eq:speed_constraints}
    \max\{|S'^\tau_m - S^\tau_m|/S^\tau_m -\epsilon_S, 0\} = 0,
\end{equation}
where $\epsilon_L$ is usually set as (smaller than) $ 10\%$.
\subsection{Constrained Primal Problem Formulation}\label{subsec:constrained}
In this subsection, we introduce the main objectives used under the untargeted setting and targeted setting.  
\paragraph{Untargeted Setting}
Under the untargeted setting, the adversary achieves its goal as long as the DNN makes a prediction other than the ground-truth label, {\em i.e.,} $\argmax_{k} \Fb_{\Thetab, k}(\xb') \neq l$. This will hold iff $F_{\Thetab, l}(\xb') < \max_{k, k\neq l}F_{\Thetab, k}(\xb')$. Therefore, we define the objective as minimizing $\max\{F_{\Thetab, l}(\xb') - max_{k, k\neq l}F_{\Thetab, k}(\xb') + conf, 0\}$, where $conf > 0$ is the desired confidence value of the DNN on the wrong prediction. Note that if the objective is equal to $0$, we have $max_{k, k\neq l}F_{\Thetab, k}(\xb') \geq F_{\Thetab, l}(\xb') + conf$.

\paragraph{Targeted Setting} 
The goal of the adversary is to render the prediction result to be the attack target $l_t$, {\em i.e.,} $\argmax_{k \in \mathcal{K}} \Fb_{\Thetab, k}(\xb') = l_t$. Therefore, the primal objective is defined as minimizing the cross entropy between $\Fb_{\Thetab, k}(\xb')$ and $l_t$, or $\max\{\max_{k, k\neq l_t}F_{\Thetab, k}(\xb') - F_{\Thetab, l_t}(\xb')  + conf, 0\}$ following the logic of the untargeted setting.

We can also adopt other objectives for our purpose. However, it turns out the above two main objectives are the most commonly-used ones in previous work \cite{kurakin2016adversarial, madry2017towards, carlini2017towards}.   
For simplicity, we denote the main loss by $\mathcal{L}(\xb, l)$. The constrained primal problem can then be formulated as 
\begin{align}
    \min\limits_{\xb'}~ & \mathcal{L}(\xb', l) \\
    \mbox{subject to}~&~\mbox{Eq.~(\ref{eq:bone_constraints}), (\ref{eq:joint_angle_constraint}), (\ref{eq:speed_constraints})} ~\nonumber
\end{align}

\begin{figure}
    \centering
    \includegraphics[width=0.48\linewidth]{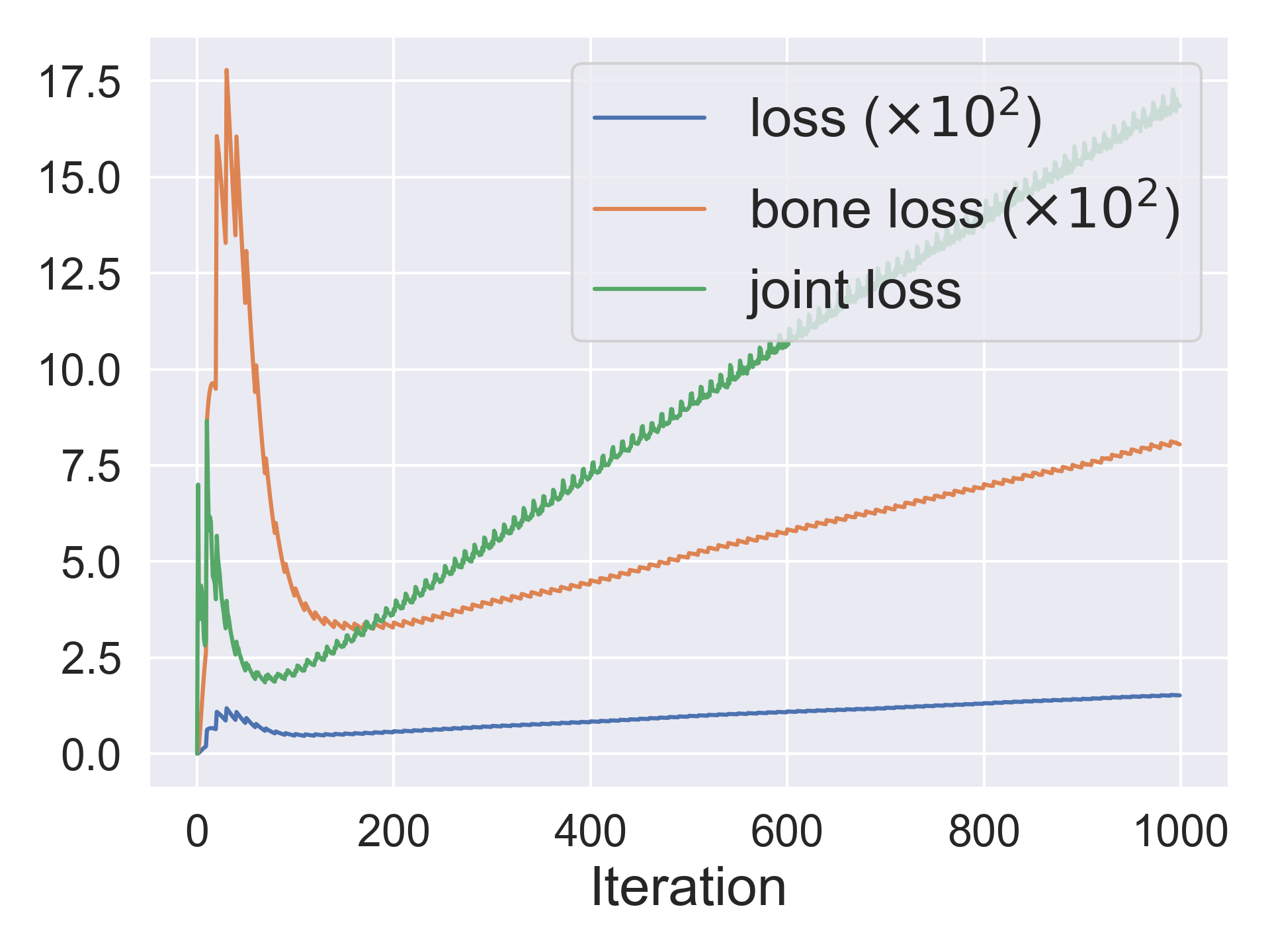}
    \includegraphics[width=0.48\linewidth]{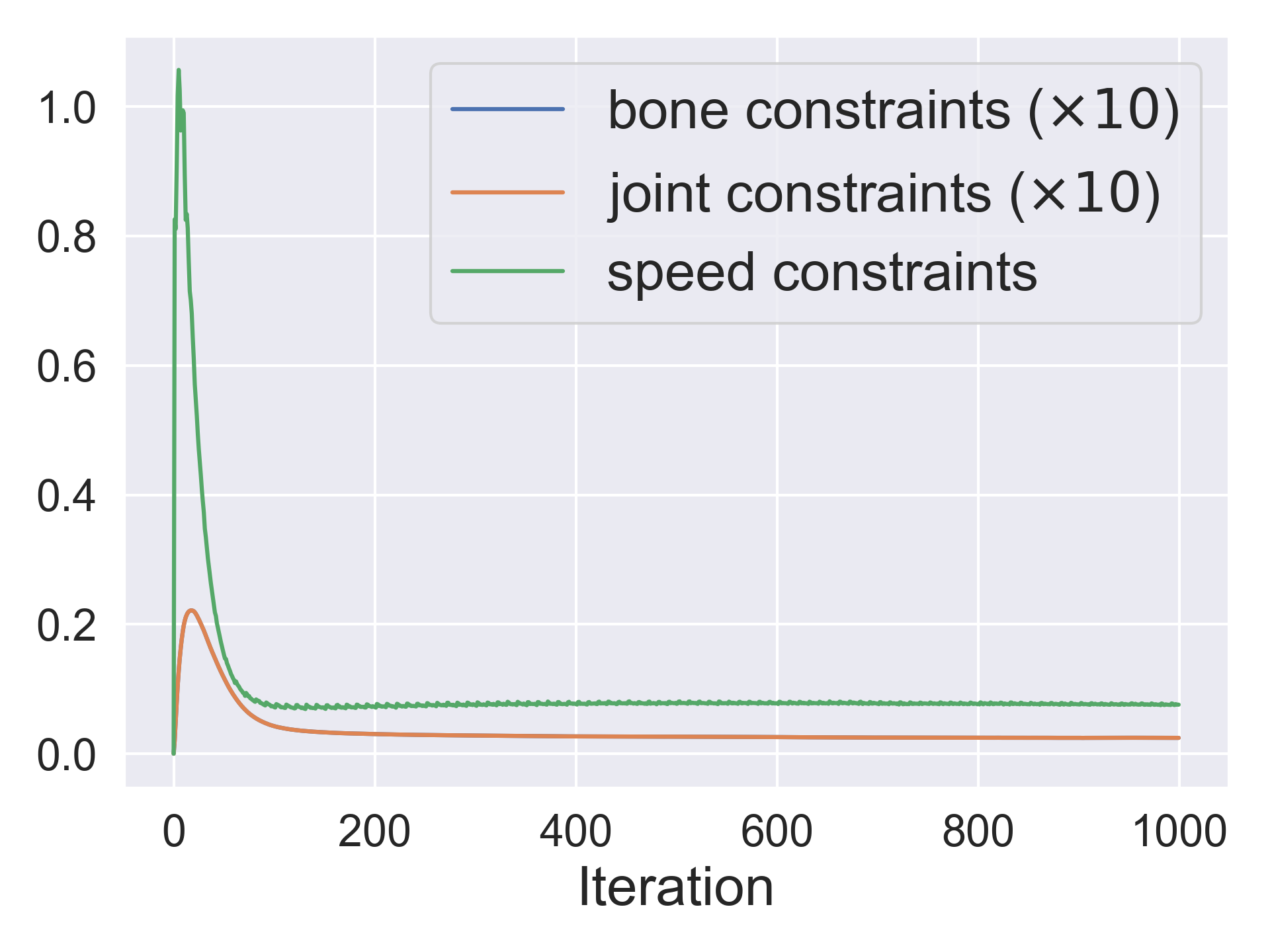}
    \vspace{-0.1cm}
    \caption{Evolution of the averaged loss items and the constraints ($\beta=1.0$)}
    \label{fig:analysis}
    \vspace{-0.1cm}
\end{figure}
\subsection{Dual Optimization by ADMM}
Note that our constrained primal problems are in general intractable. Instead of searching for a solution to the constrained primal problem, we propose to formulate and optimize its unconstrained dual problem via ADMM. The algorithm is illustrated in Alg.~\ref{alg:lagrangian}.
Specifically, we first define the augmented Lagrangian of the constrained primal as shown in Alg.~\ref{alg:lagrangian}. The additional term $\frac{\beta}{2}(\|\mathcal{\Bb}'\|_2^2 + \|\mathcal{\Jb}'\|_2^2 + \|\mathcal{\mathbf{S}}'\|_2^2)$, which is commonly used in ADMM (for nonconvex problems), aims to further penalize any violation of the equality constraints. {\em We note that larger $\beta$ usually leads to smaller violation but larger final main objective (decreases the attack success rate)}.

Specifically, given the Lagrangian $\mathcal{G}(\xb, l; \bm{\lambda}, \bm{\nu}, \bm{\omega})$ (defined in Alg.~\ref{alg:lagrangian}), the dual problem is $\max\limits_{\bm{\lambda}, \bm{\nu}, \bm{\omega}}\min\limits_{\xb'}\mathcal{G}(\xb, l; \bm{\lambda}, \bm{\nu}, \bm{\omega})$. Note that since the internal function $\min\limits_{\xb'}\mathcal{G}(\xb, l; \bm{\lambda}, \bm{\nu}, \bm{\omega})$ is an affine function w.r.t. the variables $\bm{\lambda}, \bm{\nu}, \bm{\omega}$, we can simply use single-step gradient ascent with a large step size (usually set as $\beta$ in ADMM) to update those dual variables. However, $\mathcal{G}(\xb, l; \bm{\lambda}, \bm{\nu}, \bm{\omega})$ is an extremely complicated nonconvex function w.r.t. the adversarial sample $\xb'$. Therefore, in most cases, we could only guarantee local optima for the internal minimization problem. Fortunately, it turns out that even the local optima can always fool the DNNs. 
To find a local optimum efficiently, we adopt the Adam optimizer instead of the vanilla stochastic gradient descent (SGD) since Adam optimizer always converges faster than vanilla SGD. Theoretically, a local minimum is guaranteed because the Adam optimizer stops updating the variables when the gradients are (close to) $0$. Next, we further look into the evolution of the loss during the optimization process. As shown in Fig.~\ref{fig:analysis}, at the very beginning ({\em i.e.,} the first stage), the internal minimization problem finds adversarial samples with large violation of the constraints. The large violation will cause the Lagrangian multipliers $\bm{\lambda}, \bm{\nu}, \bm{\omega}$ to increase rapidly, and thus significantly increase the loss terms $\langle \bm{\lambda}, \mathcal{\Bb}' \rangle$ (bone loss), $\langle \bm{\nu}, \mathcal{\Jb}' \rangle$ (joint loss), and $\langle \bm{\omega}, \mathcal{\mathbf{S}}' \rangle$ (speed loss). As a result, the algorithm proceeds into the second stage, where the Adam optimizer focuses more on diminishing the constraint violation $\Bb'$, $\Jb'$, and $\mathcal{\mathbf{S}}'$ when optimizing $\xb'$. Finally, the algorithm proceeds into a relatively stable stage where we can stop the algorithm. According to Fig. 2, our algorithm is very efficient in the sense that it only needs 200 (internal) iterations to enter the final stable stage.

\begin{algorithm}
	\caption{Generating Adversarial Skeleton Actions}
	\label{alg:lagrangian}
    \begin{algorithmic}
		\REQUIRE Loss function $\mathcal{L}(\xb, l)$, hyper-parameter $\beta$, adam optimizer for the adversarial skeleton action $\xb'$, maximum number of iterations $T$.
		
		\STATE \textbf{Define Constraints:} $\mathcal{B}'^\tau_j \triangleq \max\{|B'^\tau_j - B^\tau_j|/B^\tau_j - \epsilon_L, 0\}$, $\mathcal{J}'^\tau_k \triangleq \max\{J'^{\tau}_k - \epsilon_J, 0\}$, and $\mathcal{S}'^{\tau}_m \triangleq \max\{|S'^\tau_m - S^\tau_m|/S^\tau_m  -\epsilon_S, 0\}$. (Vector Representations: $\mathcal{\Bb}'$, $\mathcal{\Jb}'$, and $\mathcal{\mathbf{S}}'$)
		\vspace{0.1cm}
		\STATE \textbf{Define Lagrangian Variables:} $\bm{\lambda}$, $\bm{\nu}$, and $\bm{\omega}$ (Corresponding to $\mathcal{\Bb}'$, $\mathcal{\Jb}'$, and $\mathcal{\mathbf{S}}'$)
        \vspace{0.1cm}
		\STATE \textbf{Define Augmented Lagrangian:} $\mathcal{G}(\xb, l; \bm{\lambda}, \bm{\nu}, \bm{\omega}) \triangleq \mathcal{L}(\xb, l) + \langle \bm{\lambda},\mathcal{\Bb}' \rangle + \langle \bm{\nu}, \mathcal{\Jb}' \rangle + \langle \bm{\omega}, \mathcal{\mathbf{S}}' \rangle +  \frac{\beta}{2}(\|\mathcal{\Bb}'\|_2^2 + \|\mathcal{\Jb}'\|_2^2 + \|\mathcal{\mathbf{S}}'\|_2^2)$.
		\FOR{$t$ = $0$ to $T-1$}
		\STATE \textbf{Update $\xb'$:} fix the multipliers $\bm{\lambda}(t), \bm{\nu}(t), \bm{\omega}(t)$
		\STATE $\xb'(t+1) \in \argmin_{\xb'} \mathcal{G}(\xb', l; \bm{\lambda}(t), \bm{\nu}(t), \bm{\omega}(t))$ updated by the adam optimizer
	    \STATE \textbf{Update Multipliers:} compute $\mathcal{\Bb'}(t+1)$, $\mathcal{\Jb'}(t+1)$, and $\mathcal{\mathbf{S}'}(t+1)$ based on $\xb'(t+1)$
	    \STATE $\bm{\lambda}(t+1) = \bm{\lambda}(t) + \beta \mathcal{\Bb'}(t+1)$; $\bm{\nu}(t+1) = \bm{\nu}(t) + \beta \mathcal{\Jb'}(t+1)$; $\bm{\omega}(t+1) = \bm{\omega}(t) + \beta \mathcal{\mathbf{S}'}(t+1)$
	
		\ENDFOR
		\STATE \textbf{Output} $\xb'(T)$
	\end{algorithmic}
\end{algorithm}

\begin{figure*}
    \centering
    \includegraphics[width=0.9\linewidth]{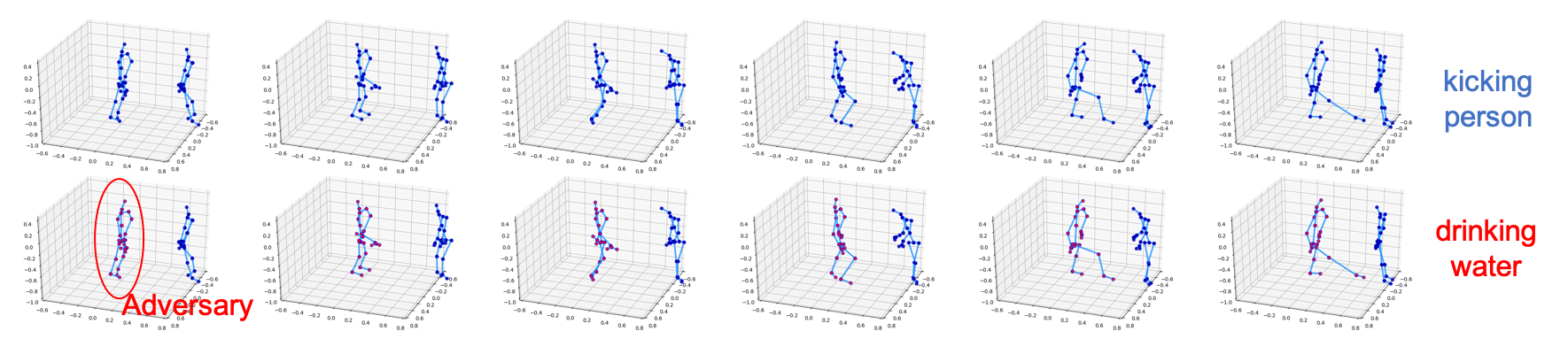}
    \vspace{-0.2cm}
    \caption{The top six frames represent a ``kicking (another person)" skeleton action, and the bottom six frames are the corresponding frames from the adversarial skeleton action generated by our attack under the targeted setting (optimizing the first person). The generated adversarial skeleton action is recognized as ``drinking water" by the 2s-AGCN.}
    \label{fig:visual}
\end{figure*}
\section{Defense against Adversarial Skeleton Actions}
Note that although the method proposed in \cite{li2018second, cohen2019certified} can certify larger robust radii than \cite{lecuyer2018certified}. However, the sample complexity to compute the confidence intervals in \cite{li2018second, cohen2019certified} will lead to computational overhead in the inference stage. {\em Therefore, we only use the method in \cite{cohen2019certified} in the certification process. In the inference stage, we modify the method in \cite{lecuyer2018certified} to build a relatively efficient defense against adversarial skeleton actions, as shown in Alg.~\ref{alg:randomized_defense}}. In general, our proposed defense consists of two steps: adding Gaussian noise and temporal filtering by Gaussian kernel. In the following, we will detail these two steps and explain why we include them in the defense. 
\subsection{Additive Gaussian Noise}
Our first step is adding Gaussian noise to the skeleton actions. 
In the inference stage, we follow \cite{lecuyer2018certified} to make the prediction as $\argmax_k E(\Fb_{\Thetab, k}(\mathcal{M}(\xb')))$ given input $\xb'$, where $\mathcal{M}(\xb) = \Gb(\xb + \zb)$ is randomized mechanism with Gaussian noise $\zb$ and post-processing function $\Gb$. 
In order to estimate $E(\Fb_{\Thetab}(\mathcal{M}(\xb')))$, we sample N noisy samples $\tilde\xb'(n) = \xb'+\tilde\zb(n)$ from $\mathcal{N}(\xb', \sigma^2\Ib)$ and feed them into the post-processing function $\Gb$ and the neural network $\Fb_\Thetab$. $E(\Fb_{\Thetab}(\mathcal{M}(\xb')))$ is estimated by $\frac{1}{N}\sum_{n=1}^{N}\Fb_{\Theta}(\Gb(\tilde\xb'(n)))$, and according to the Chernoff bound \cite{boucheron2013concentration}, the error of this estimation is bounded by $$
Pr(|\frac{1}{N}\sum_{n=1}^{N}\Fb_{\Theta, l}(\Gb(\tilde\xb'(n))) - E(\Fb_{\Thetab, l}(\mathcal{M}(\xb')))| < \epsilon) \sim \mathcal{O}(e^{-Nt^2})
$$.  

In the certification stage, we rely on the main theorem from \cite{cohen2019certified}, which gives the currently tightest bound:
\begin{lemma}\label{lem:lemma2}
Denote an mechanism randomized by Gaussian noise by $\mathcal{M}(\xb) = \Gb(\xb + \zb)$, and the ground-truth label by $l$. Define $f(\xb) = \argmax_k\Fb_{\Thetab, k}(\mathcal{M}(\xb))$. Suppose $\underline{p_A}$ \& $\overline{p_B}$ satisfy
\begin{align}
    Pr(f(\xb) = l)  \geq \underline{p_A} \geq \overline{p_B} \geq  \max_{i\neq l} Pr(f(\xb) = i),
\end{align}
the $\ell_2$ robust radius is $R = \frac{\sigma}{2}(\Phi^{-1}(\underline{p_A}) - \Phi^{-1}(\overline{p_B}))$.
\end{lemma}
Lemma~\ref{lem:lemma2} indicates that as long as $\|\xb' - \xb\|_2 < R$, $\argmax_i Pr(f(\xb) = i) = l$, {\it i.e.}, the prediction is correct. The algorithm using the above lemma for certification is detailed in Algorithm~\ref{alg:randomized_certification}. In the next subsection, we will detail the post-processing function mentioned before.
\begin{algorithm}
	\caption{Defense (Inference)}
	\label{alg:randomized_defense}
	\begin{algorithmic}
		\REQUIRE Neural Network $\Fb_\Thetab(\cdot)$, standard deviation of the additive Gaussian noise $\sigma$, skeleton action $\xb'$ (probably adversarial), number of noisy samples for inference of $n$.
		
		\STATE Sample $N$ samples from $\mathcal{N}(\xb', \sigma^2\Ib)$ $\rightarrow$ $\{\tilde\xb'(n)|n=1, 2,..., N\}$
		\vspace{0.1cm}
		\STATE Smooth $\tilde\xb'(n)$ by a $1\times5$ or $1\times7$ Gaussian filter $\rightarrow$ $\Gb(\tilde\xb'(n))$
        \vspace{0.1cm}
		\STATE Feed $\Gb(\tilde\xb'(n))$ into the network $\rightarrow$ $\Fb_\Theta(\Gb(\tilde\xb'(n)))$
		\STATE \textbf{Output} $\argmax_l\sum_{n=1}^{N}\Fb_{\Theta, l}(\Gb(\tilde\xb'(n)))$
	\end{algorithmic}
\end{algorithm}

\begin{algorithm}
	\caption{Defense (Certification)}
	\label{alg:randomized_certification}
	\begin{algorithmic}
		\REQUIRE Neural Network $\Fb_\Thetab(\cdot)$, standard deviation of the additive Gaussian noise $\sigma$, original and adversarial skeleton action $\xb$ \& $\xb'$, number of noisy samples for inference of $n$, a predefined confidence value p for hypothesis test (usually $95\%$).
		
		\STATE \textbf{Recognition:} Sample $N$ samples from $\mathcal{N}(\xb, \sigma^2\Ib)$ $\rightarrow$ $\{\tilde\xb(n)|n=1, 2,..., N\}$
		\vspace{0.1cm}
		\STATE Smooth $\tilde\xb(n)$ by a $1\times5$ or $1\times7$ Gaussian filter $\rightarrow$ $\Gb(\tilde\xb(n))$
        \vspace{0.1cm}
		\STATE Feed $\tilde\xb(n)$ into the (normally trained) network $\rightarrow$ $\Fb_\Theta(\Gb(\tilde\xb(n)))$
		\STATE \textbf{Confidence Interval:} Compute the number (counts) top two indices in $\{\argmax_k\Fb_{\Theta, k}(\Gb(\tilde\xb(n)))~|~n=1,2,...,N\}$ $\rightarrow c_A, c_B$
		\STATE Compute the lower bound for $p_A$ and the upper bound for $p_B$ by the method in \cite{goodman1965simultaneous} with confidence $p$ $\rightarrow \underline{p_A}, \overline{p_B}$.
		\STATE \textbf{Certification:} Compute the certified $\ell_2$ radius by $R = \frac{\sigma}{2}(\Phi^{-1}(\underline{p_A}) - \Phi^{-1}(\overline{p_B}))$.

        \STATE Output $\max\{R, 0\}$ \textbf{if} {\em $p_A$ corresponds to the ground-truth label $l$} \textbf{else} $-1$ (certified robust radius)
		\STATE Compare R with $\|\xb'-\xb\|_2$, and if $R$ is larger, then output the index corresponding to $c_A$ (if inputs have $\xb'$)
	\end{algorithmic}
\end{algorithm}

\subsection{Temporal Filtering by Gaussian Kernel}
After adding Gaussian noise to the skeleton actions, we propose to further smooths the action along the temporal axis by a $1\times5$ or $1\times7$ Gaussian filter. The intuition is that the adjacent frames in a skeleton action sequence are very similar to each other, and thus can be used as references to rectify the adversarial perturbations. 
Although this additional operation does not improve the certification results, we observe that it can help our defense become more compatible with a normally trained model than the original randomized smoothing method in \cite{lecuyer2018certified, cohen2019certified}.
Also, we argue that this simple operation is not usually used in previous work because it is not very suitable in the image recognition domain, where no adjacency information (along the temporal axis) is available. 

\begin{table*}
    \begin{center}
    \vspace{0.0cm}
    \scalebox{0.95}{
    \begin{tabular}{|c|c|ccccc|ccccc|}
    \hline
        \textbf{White-box} & \multirow{2}{*}{$\beta$} & \multicolumn{5}{c|}{\textbf{NTU CV}} & \multicolumn{5}{c|}{\textbf{NTU CS}}\\
         \textbf{Untargeted} & & Success Rate & $\Delta B/B$ & $\Delta J$ & $\Delta K/K$ & $\ell_2$ & Success Rate & $\Delta L/L$ & $\Delta J$ & $\Delta K/K$ & $\ell_2$\\
         \hline
        \multirow{3}{*}{\textbf{HCN}}
        & $0.1$ & 100\% & 2.64\% & 0.132 & 4.52\% & 0.396 & 100\%  & 2.17\% & 0.111  & 3.17\% & 0.347\\
        & $1.0$  & 100\% & 1.92\% & 0.099 & 1.65\% & 0.330 &  100\%  &1.62\% & 0.086 & 1.30\% & 0.290\\
        & $10.0$ & 92.8\% & 1.50\% & 0.085 & 1.25\% & 0.270 &  92.4\%  & 1.25\% & 0.073 & 0.98\% & 0.241\\
        \hline \hline
        \multirow{3}{*}{\textbf{2s-AGCN}}
        & $0.1$ & 100\% & 2.17\% & 0.112 & 1.62\% & 0.653 & 100\%  & 1.97\% & 0.107  & 2.20\% & 0.614\\
        & $1.0$  & 100\% & 1.70\% & 0.094 & 0.59\% & 0.528 & 100\%  & 1.46\% & 0.086 & 0.57\% & 0.496\\
        & $10.0$  & 99.0\% & 1.37\% & 0.083 & 0.39\% & 0.428 & 98.8\%  & 1.19\% & 0.078 & 0.34\% & 0.413\\
        \hline 
        \hline
        \textbf{White-box} & \multirow{2}{*}{$\beta$} & \multicolumn{5}{c|}{\textbf{NTU CV}} & \multicolumn{5}{c|}{\textbf{NTU CS}}\\
         \textbf{targeted} & & Success Rate & $\Delta B/B$ & $\Delta J$ & $\Delta K/K$ & $\ell_2$ & Success Rate & $\Delta L/L$ & $\Delta J$ & $\Delta K/K$ & $\ell_2$\\
         \hline
        \multirow{3}{*}{\textbf{HCN}}
        & $0.1$ & 100\% & 3.60\% & 0.165 & 7.75\% & 0.673 & 100\% & 3.55\% & 0.165  & 6.68\% & 0.723\\
        & $1.0$  & 99.7\% & 3.24\% & 0.156 & 4.69\% & 0.630 & 100\%  & 3.16\% & 0.155 & 4.24\% & 0.674\\
        & $10.0$ & 22.3\% & 2.27\% & 0.115 & 2.83\% & 0.444 & 26.9\%  & 2.14\% & 0.112 & 2.50\% & 0.462\\
        \hline \hline
        \multirow{3}{*}{\textbf{2s-AGCN}}
        & $0.1$ & 100\% & 1.66\% & 0.090 & 0.55\% & 0.569& 100\%  & 1.67\% & 0.091  & 0.71\% & 0.649\\
        & $1.0$   & 100\% & 1.61\% & 0.091 & 0.42\% & 0.556 & 100\%  & 1.56\% & 0.090 & 0.49\% & 0.615\\
        & $10.0$   & 97.2\% & 1.54\% & 0.089 & 0.38\% & 0.512 &  97.9\%  & 1.47\% & 0.087 & 0.40\% & 0.552\\
        \hline
    \end{tabular}}
    \vspace{0.1cm}
    \caption{The {\em empirical} performance of our proposed method: averaged bone-length difference between original and adversarial skeletons ($\Delta L/L$), averaged joint angle difference (upper bound) ($\Delta J$), kinetic energy difference ($\Delta K/K$), $\ell_2$ distance ($\ell_2$).}\label{tab:attack_performance}
    \end{center}
    \vspace{-0.4cm}
\end{table*}
\section{Experiments}
\subsection{Attack Performance}\label{sec:attack_performance}
\paragraph{Main Results}
The main results of our attack are shown in Table~\ref{tab:attack_performance}. As we can see, our proposed attack can achieve 100\% success rates with very small violation of the constraints. 
The averaged normalized bone-length difference is approximately $1\% \sim 2\%$, and the violation of the joint angles is smaller than $10^\circ$. {\em Considering the skeleton data is usually noisy, this subtle violation is considered ``very common" in real world.} 

We also note that adversarial-sample generation under the untargeted setting is usually easier than that under the targeted setting since a targeted adversarial sample is guaranteed to be an untargeted adversarial sample, but not vice versa. This fact is also reflected by the results in Table~\ref{tab:attack_performance}. Furthermore, in Fig.~\ref{fig:visual}, we show the visualization result of an adversarial skeleton action (recognized as a normal action ``drinking water") generated by our attack, which is almost visually indistinguishable from its original skeleton action (``kicking").
\begin{table}
    \begin{center}
    \scalebox{0.95}{
    \begin{tabular}{c|c|cc}
        \textbf{Source (Model) $\rightarrow$ Target} & \textbf{Dataset} & $\beta=0.1$ & $\beta=0.01$ \\
        \hline \hline
        \multirow{2}{*}{\textbf{HCN(1) $\rightarrow$ HCN(2)}} & NTU CV & 24.7\% & 26.0\% \\
         & NTU CS & 28.5\% & 32.6\% \\
        \hline \hline
        \multirow{2}{*}{\textbf{HCN(1) $\rightarrow$ 2s-AGCN}} & NTU CV & 16.2\% & 20.4\% \\
         & NTU CS & 17.6\% & 21.8\% \\
    \end{tabular}}
    \caption{Attack success rates of adversarial examples transferred between models.}
    \vspace{-0.4cm}
    \label{tab:trans}
    \end{center}
\end{table}
\paragraph{Transferability} In order to shed light on the transferability of our attack, we feed the adversarial skeleton actions generated on a HCN model to another HCN model and 2s-AGCN, respectively. In order to boost the transferability performance, we set $\beta$ as $0.01$ or $0.1$ to generate adversarial skeleton actions with larger perturbation. The attack success rates are given in Table~\ref{tab:trans}. {\em Similar to 3D adversarial point clouds \cite{xiang2018generating}, the transferability of the adversarial skeleton actions is also a little limited compared with adversarial images.}
\paragraph{Comparison with C\&W Attack}
We use C\&W attack as an example to shed light on the difference between our attack and the existing attacks. C\&W attack has been demonstrated as a successful optimization-based adversarial attack in many application domains. However, since C\&W attack mainly considers minimizing the $\ell_2$ distance between original and adversarial skeletons, it might easily violate the constraints, as shown in our simple case study (Table~\ref{tab:comparison}).
\begin{table}
    \begin{center}
    \scalebox{0.95}{
    \begin{tabular}{c|ccccc}
    \hline
       \textbf{Untargeted} 
        & Success Rate & $\Delta B/B$ & $\Delta J$ & $\Delta K/K$ & $\ell_2$\\
        \hline
         NTU CV & 100\% & 4.67\% & 0.241 & 13.0\% & 0.278\\
        \hline
         NTU CS & 100\% & 4.09\% & 0.211 & 10.2\% & 0.244\\
         \hline
         \hline
        \textbf{Targeted} 
        & Success Rate & $\Delta B/B$ & $\Delta J$ & $\Delta K/K$ & $\ell_2$\\
        \hline
         NTU CV & 100\% & 8.82\% & 0.468 & 38.1\% & 0.510\\
        \hline
         NTU CS & 100\% & 9.45\% & 0.507 & 36.8\% & 0.520\\
    \hline
    \end{tabular}}
    \caption{Adversarial skeleton actions generated by C\&W attack on HCN.}\label{tab:comparison}
    \end{center}
    \vspace{-0.3cm}
\end{table}

\subsection{Defense Performance}
\begin{figure}
    \centering
    \includegraphics[width=0.48\linewidth]{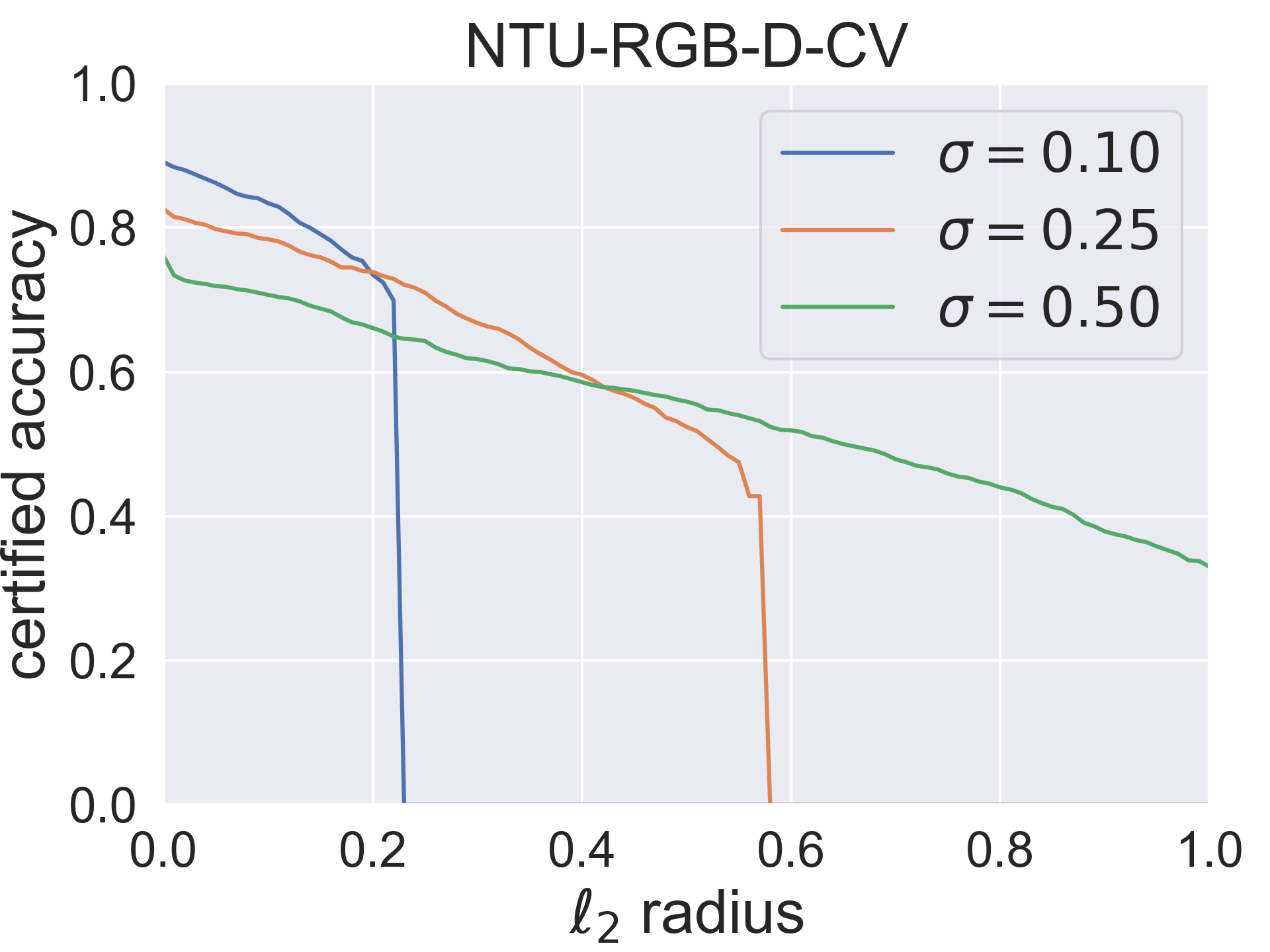}
    \includegraphics[width=0.48\linewidth]{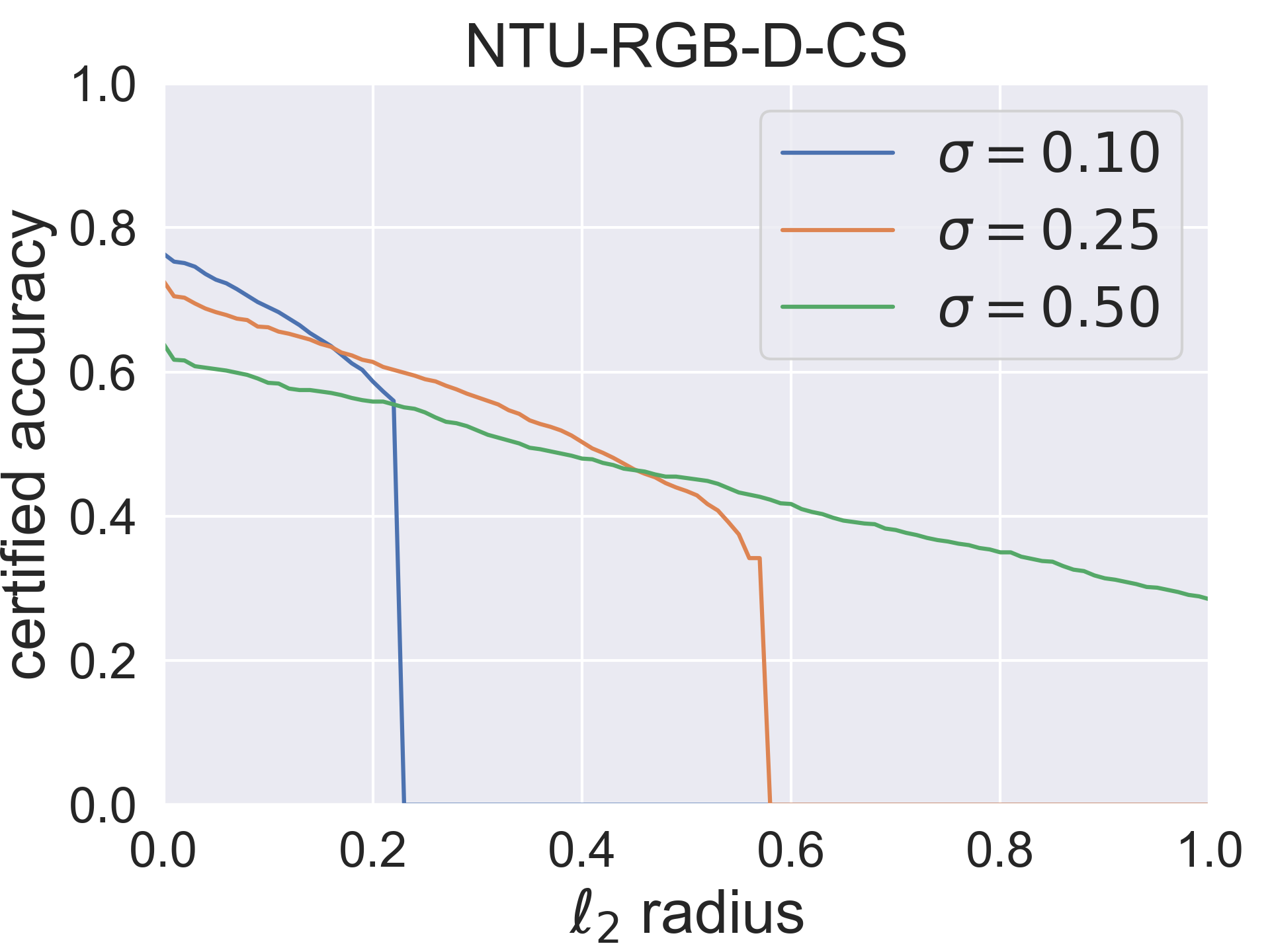}
    \vspace{-0.1cm}
    \caption{Certification accuracy on HCN}
    \label{fig:certify}
    \vspace{-0.1cm}
\end{figure}

\paragraph{Empirical Results} 
We demonstrate the performance of the defense for inference in Table~\ref{tab:empirical_defense}. We set $\beta=1.0$ to generate adversarial examples, and set $N=50$ (Alg.~\ref{alg:randomized_defense}), which is more smaller than the number of samples required for certification but can achieve good empirical performance, as shown in Table~\ref{tab:empirical_defense}. {\em it is much easier to defend adversarial skeleton actions under the targeted setting than the untargeted setting.} Note that the accuracy of HCN on NTU-CV and NTU-CS is respectively $91.1\%$ and $86.5\%$ \cite{li2018co}, and the accuracy of 2s-AGCN is respectively $95.1\%$ and $88.5\%$ \cite{shi2019two}. 
\begin{table}
    \begin{center}
    \vspace{0.0cm}
    \scalebox{0.9}{
    \begin{tabular}{c|c|cc|cc}
        \multirow{2}{*}{Model} & \multirow{2}{*}{Setting} & \multicolumn{2}{c|}{\textbf{NTU CV}} & \multicolumn{2}{c}{\textbf{NTU CS}} \\
        &  & $\sigma=0.01$ & $\sigma=0.02$ & $\sigma=0.01$ & $\sigma=0.02$ \\
        \hline
        \hline
        \multirow{2}{*}{HCN} & Untargeted & 62.0\% & 62.3\% & 50.6\% & 51.4\% \\
                             & Targeted & 79.4\% & 70.8\% & 67.1\% & 58.3\%\\
        \hline
        \multirow{2}{*}{2s-AGCN} & Untargeted & 51.0\% & 42.2\% & 42.1\% & 40.2\%\\
        & Targeted & 60.8\% & 50.5\% & 42.2\% & 44.1\% \\
    \end{tabular}}
    \vspace{0.1cm}
    \caption{{\em Empirical} performance (model accuracy) of our proposed defense on normally trained models. 
    }\label{tab:empirical_defense}
    \end{center}
\end{table}
\paragraph{Certified Results} Due to the high computational cost of the certification method (N=1000), we mainly evaluate the certification algorithm on HCN. The certified accuracy achieved by different levels of noise is shown in Fig.~\ref{fig:certify}. Note that we use the same level of noise to train the model as the noise for certification. 
{\em As we can see, with sacrificing $10\%\sim20\%$ accuracy on the clean samples, the method is able to achieve about $50\%$ certified accuracy ($\ell_2 = 0.5$)}.

\subsection{Additional Experimental Results}
\begin{figure*}[h]
    \centering
    \includegraphics[width=0.9\linewidth]{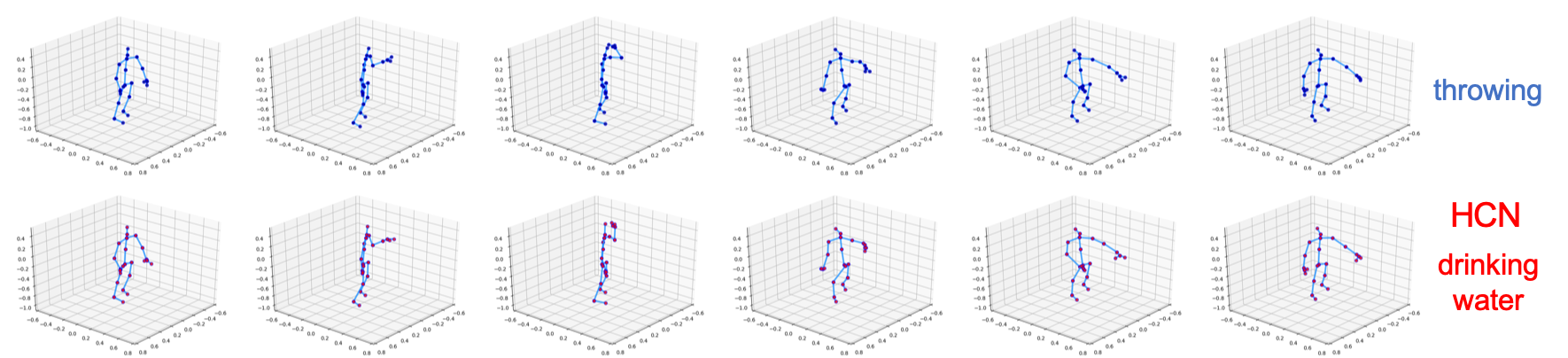}
     \includegraphics[width=0.9\linewidth]{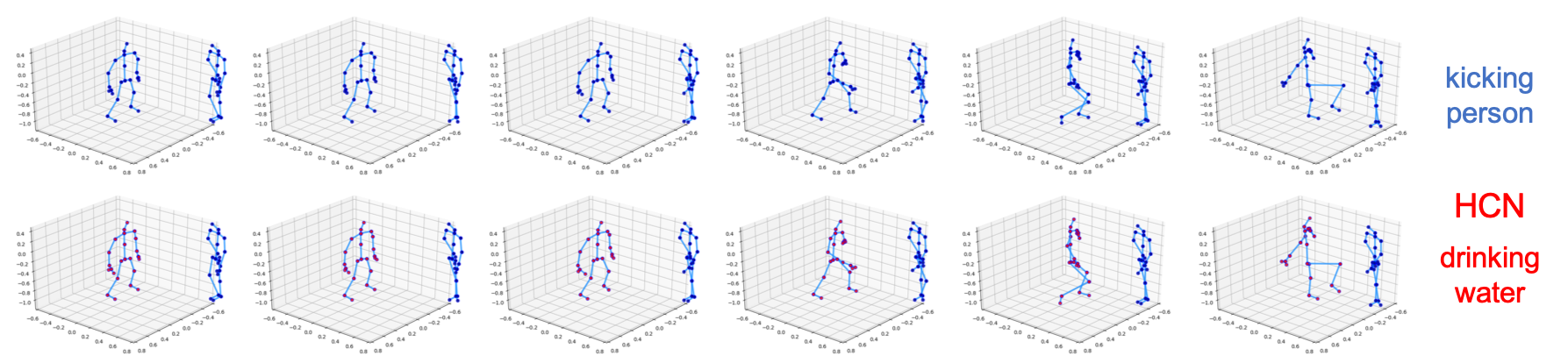}
    \caption{The adversarial skeleton actions generated by our attack under the targeted setting. The generated adversarial skeleton actions are recognized as ``drinking water" by the HCN.}
    \label{fig:visual_hcn}
\end{figure*}

\begin{figure*}
    \centering
    \includegraphics[width=0.9\linewidth]{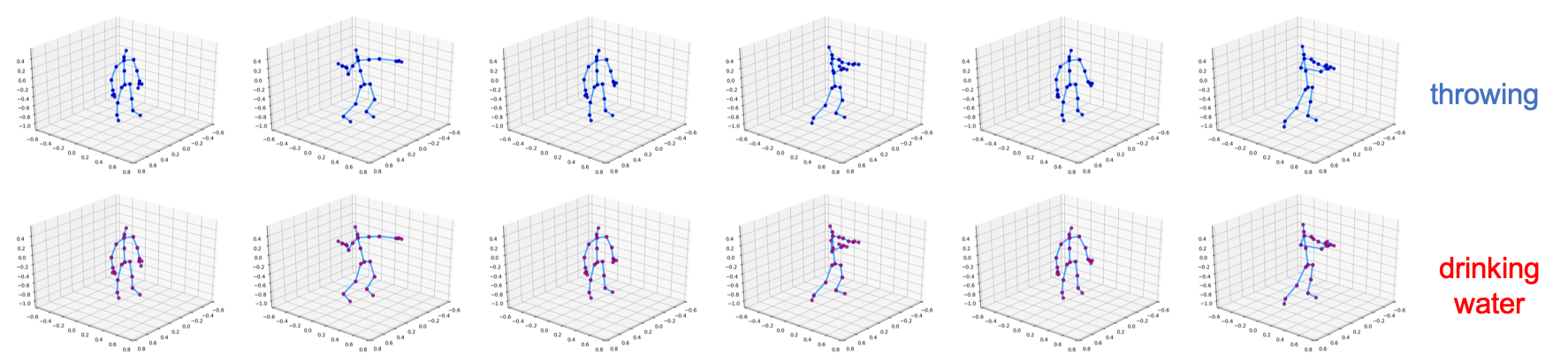}
    \includegraphics[width=0.9\linewidth]{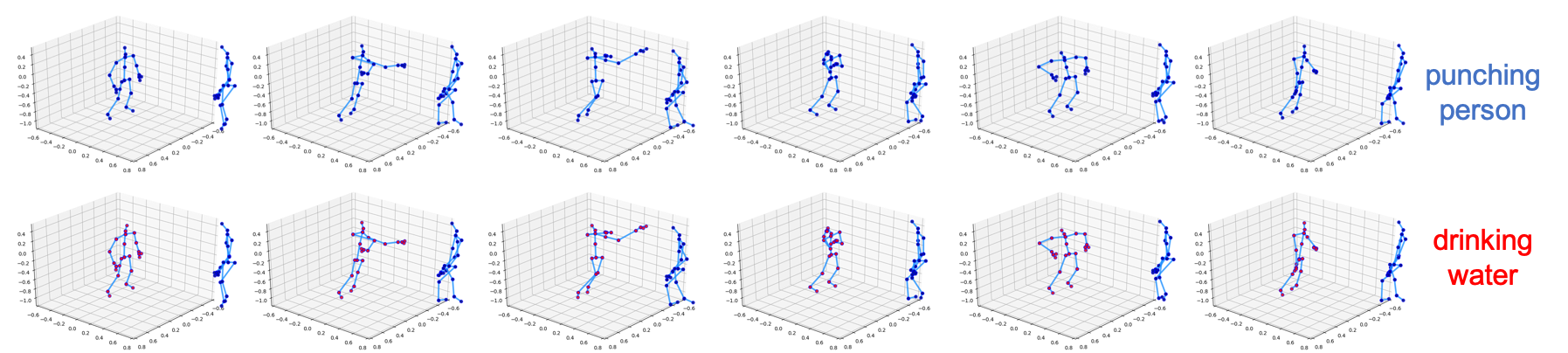}
    \caption{The adversarial skeleton actions generated by our attack under the targeted setting. The generated adversarial skeleton actions are recognized as ``drinking water" by the 2s-AGCN.}
    \label{fig:visual_agcn}
\end{figure*}

\paragraph{Additional visualization results}
Here we provide more visualization results. We use ``drinking water" as the attack target because ``drinking water" is a normal action, which looks completely different from the some violent/abnormal actions like throwing, kicking, pushing, and punching. Despite the obvious visual differences between ``drinking water'' and those abnormal actions, our attack can still fool the state-of-the-art models to recognize those abnormal actions as ``drinking water'' by imperceptible and reproducible perturbation.
In Fig.~\ref{fig:visual_hcn}, we show that our attack can fool the HCN model to recognize the ``throwing" and ``kicking a person'' actions as a normal action ``drinking water" by imperceptible adversarial perturbation. Similarly, in Fig.~\ref{fig:visual_agcn}, we show that our attack can fool the 2s-AGCN model to recognize the ``throwing'' and ``punching a person" actions as a normal action ``drinking water".
{\em We also attach the more videos to show the original and adversarial skeleton actions in the supplementary material}. These visualization results along with the quantitative results in Table 1 (in the paper) demonstrate that the perturbations are indeed imperceptible and reproducible.

\paragraph{Kinetics Dataset} Except for the NTU dataset, we also evaluate our attack on another popular dataset, {\em i.e.,} Kinetics-400 dataset under both the untargeted and targeted settings.
As shown in Table~\ref{tab:attack_performance}, under the untargeted setting, our attack can achieve 100\% attack success rates with very small violation of the constraints, similar to its performance on the NTU dataset. However, under the targeted setting, it is much more difficult for our attack to find targeted adversarial skeleton actions with very small violations of the constraints. This is because Kinetics-400 has 400 classes of actions, and the original NTU dataset only has 60 classes of actions. Also, we argue that the results on Kinetics under the targeted setting do not devalue our attack since, even for most of the clean testing samples from Kinetics, it is difficult for the state-of-the-models to predict their ground-truth labels (targets). 

\begin{table}
    \begin{center}
    \vspace{0.0cm}
    \scalebox{0.9}{
    \begin{tabular}{|c|c|ccccc|}
    \hline
        \multirow{2}{*}{\textbf{Untargeted}} & \multirow{2}{*}{$\beta$} & \multicolumn{5}{c|}{\textbf{Kinetics-400}}\\
          & & Success Rate & $\Delta B/B$ & $\Delta J$ & $\Delta K/K$ & $\ell_2$\\
         \hline
        \multirow{3}{*}{\textbf{HCN}}
        & $0.1$ & 100\% & 2.60\% & 0.082 & 1.66\% &0.150\\
        & $1.0$  & 100\% & 2.58\% & 0.080 & 1.52\% & 0.162 \\
        & $10.0$ & 98.8\% & 2.49\% & 0.078 & 1.21\% & 0.145\\
        \hline \hline
        \multirow{3}{*}{\textbf{2s-AGCN}}
        & $0.1$ & 100\% & 0.91\% & 0.053 & 0.58\% & 0.331 \\
        & $1.0$  & 100\% & 0.77\% & 0.047 & 0.53\% & 0.298 \\
        & $10.0$  & 100\% & 0.75\% &  0.046 & 0.52\% & 0.287\\
        \hline
        \hline
        \multirow{2}{*}{\textbf{Targeted}} & \multirow{2}{*}{$\beta$} & \multicolumn{5}{c|}{\textbf{Kinetics-400}}\\
          & & Success Rate & $\Delta B/B$ & $\Delta J$ & $\Delta K/K$ & $\ell_2$\\
        \hline
        \multirow{3}{*}{\textbf{HCN}}
        & $0.1$ & 90.2\% & 5.22\% & 0.220 & 11.2\% & 1.864\\
        & $1.0$  & 67.2\% & 2.79\% & 0.124 & 4.86\% & 1.350 \\
        & $10.0$ & 17.2\% & 1.44\% & 0.073 & 2.36\% & 0.763\\
        \hline \hline
        \multirow{3}{*}{\textbf{2s-AGCN}}
        & $0.1$ & 99.2\% & 5.25\% & 0.167 & 1.20\% & 0.725 \\
        & $1.0$  & 98.8\% & 5.04\% & 0.159 & 1.21\% & 0.722 \\
        & $10.0$  & 98.4\% & 4.89\% &  0.153 & 1.03\% & 0.677\\
        \hline
    \end{tabular}}
    \vspace{0.2cm}
    \caption{The performance of our proposed attack on Kinetics: success rate, averaged bone-length difference between original and adversarial skeletons ($\Delta L/L$), averaged joint angle difference (upper bound) ($\Delta J/J$), kinetic energy difference ($\Delta K/K$), $\ell_2$ distance ($\ell_2$).}\label{tab:attack_performance}
    \end{center}
\end{table}

\section{Conclusion}
We study the problem of adversarial vulnerability of skeleton-based action recognition. 
We first identify and formulate three main constraints that should be considered in adversarial skeleton actions. 
Since the corresponding constrained optimization problem is intractable, we propose to optimize its dual problem by ADMM, which is a generic method proposed in this paper to generate adversarial examples with complicated constraints. 
To defend against adversarial skeleton actions, we further specify an efficient defensive inference algorithm and a certification algorithm. The effectiveness of the attack and defense is demonstrated on two opensource models, and the results induce several interesting observations that can help us understand the properties of adversarial skeleton actions.

\bibliographystyle{ACM-Reference-Format}
\bibliography{sample-base}

\appendix

\end{document}